\documentclass{article}

\usepackage{arxiv}

\usepackage[utf8]{inputenc} 
\usepackage[T1]{fontenc}    
\usepackage{hyperref}       
\usepackage{url}            
\usepackage{booktabs}       
\usepackage{amsfonts}       
\usepackage{nicefrac}       
\usepackage{microtype}      
\usepackage{lipsum}
\usepackage{graphicx}
\usepackage[round, longnamesfirst]{natbib} 
\usepackage{caption}
\usepackage{subcaption}
\usepackage{diagbox}
\usepackage{booktabs}
\usepackage[longend]{algorithm2e}
\usepackage{textgreek}
\usepackage{tcolorbox}

\usepackage{amsmath}
\usepackage{amssymb}
\usepackage{graphicx}
\DeclareMathOperator*{\argmax}{argmax}
\usepackage{setspace}

\title{Sample Efficiency in Sparse Reinforcement Learning: Or Your Money Back}

\author{
  Trevor A. McInroe \\
  Northwestern University\\
  \texttt{trevormcinroe2022@u.northwestern.edu} \\
}

\begin{document}
\maketitle

\begin{abstract}
Sparse rewards present a difficult problem in reinforcement learning and may be inevitable in certain domains with complex dynamics such as real-world robotics. Hindsight Experience Replay (HER) is a recent replay memory development that allows agents to learn in sparse settings by altering memories to show them as successful even though they may not be. While, empirically, HER has shown some success, it does not provide guarantees around the makeup of samples drawn from an agent's replay memory. This may result in minibatches that contain only memories with zero-valued rewards or agents learning an undesirable policy that completes HER-adjusted goals instead of the actual goal.

In this paper, we introduce \textit{Or Your Money Back} (OYMB), a replay memory sampler designed to work with HER. OYMB improves training efficiency in sparse settings by providing a direct interface to the agent's replay memory that allows for control over minibatch makeup, as well as a preferential lookup scheme that prioritizes real-goal memories before HER-adjusted memories. We test our approach on five tasks across three unique environments. Our results show that using HER in combination with OYMB outperforms using HER alone and leads to agents that learn to complete the real goal more quickly.
\end{abstract}


\section{Introduction}
Reinforcement learning (RL) has shown great success in virtual environments, where generating a large number of agent-environment interactions and experimenting with various reward functions is feasible in a short amount of time. However, the application of RL to real-world systems, such as robotics, is limited by the challenges of relatively slow data collection and the difficulty in properly specifying a reward function~\citep{komer_2013}. The former motivates research with the aim of improving learning efficiency in reinforcement learning agents~\citep{YangDujia2020SERL, WenZheng2017ERLi}. The latter has led to a parallel branch of research that focuses on algorithms capable of handling simple, sparse reward functions~\citep{SeoMinah2019RPCA, RenHailin2020Arlt}. 

To overcome these issues in tandem,~\citet{andry_2017} introduced Hindsight Experience Replay (HER), which is specifically designed for sparse-reward settings. HER works by artificially creating positive reward experiences. After each episode, regardless of the real goal, the agent's memory is adjusted to show the episode's trajectory as successful. While HER makes it possible for agents to learn in sparse-reward settings, it provides no guarantees around replay memory sampling behavior. This may lead to a large number of samples containing only non-informative experiences or to agents that learn policies to complete HER-adjusted goals instead of the actual goal.

In this paper, we introduce \textit{Or Your Money Back} (OYMB), a novel sampler for the HER replay memory that allows for faster, more stable convergence of reinforcement learning agents in sparse-reward settings. Our method builds off of the success of HER by constructing an interface to the agent's replay memory that provides guarantees around experience tuple selection during agent training. Also, it provides a preferential-lookup scheme that selects real-goal-achieving memories before HER-altered memories, thus discouraging the agent from learning policies that achieve the incorrect goal.

\section{Background and Related Works}
\subsection{Reinforcement Learning}
Reinforcement learning can be framed as a sequential decision-making problem represented by a finite Markov decision process. An agent interacts with an environment, $\mathcal{E}$, over a series of time steps, $t$\footnote{For notational brevity, current timesteps, $x$, have no special marking, and steps one into the future are marked $x^{\prime}$}, over the length of an episode, $T$. At each step, the agent observes a representation of the environment's state, $s \in \mathcal{S}$, and from that observation chooses an action, $a \in \mathcal{A}$. Based on the merit of the agent's action choice, a reward function produces a scalar feedback signal, $\mathcal{R}: \mathcal{S} \times \mathcal{A} \rightarrow \mathbb{R}$. The agent's purpose is to identify an action-selection policy, $\pi: \mathcal{S} \rightarrow \mathcal{A}$, that maximizes its discounted cumulative reward over the lifetime of a given task. From these dynamics, we arrive at the following iterative \textit{Bellman equation}, that the agent updates through its interactions: 
\begin{gather*}
    Q^{\pi}(s,a) = \mathbb{E}[r + \gamma \argmax_{a^{\prime}} Q^{\pi}(s^{\prime},a^{\prime})]
\end{gather*}
where any arbitrary \textit{Q-function} converges to optimality $Q^{\pi} \rightarrow Q^*$ as $t \rightarrow \infty$~\citep{sutton_barto_2018}. The Q-function can be any parameterized function, either linear or nonlinear. 

\subsection{Deep Q-Networks}
Deep Q-Networks (DQN), introduced by~\citet{dqn}, is a model-free, deep RL algorithm designed to work in discrete action settings. Given a neural network, $\theta$, a DQN is trained by minimizing the convex loss $\mathcal{L}_i(\theta_i) = [y_i - Q(s,a;\theta_i)]^2$. Unlike standard supervised learning models, we do not have an explicit ``ground truth'' to use for minimizing this loss. Instead, we use the temporal differece (TD) target, i.e., the reward and Q-values for the next state: $y_i = r + \gamma \argmax_{a^{\prime}}Q(s^{\prime},a^{\prime};\theta_{i-N})$. Note that the authors use the network weights from a previous iteration, $\theta_{i-N}$, to help add stability to training. Together, this presents the following gradient over a sample of experiences:
\begin{gather*}
    \nabla_{\theta_i}\mathcal{L}_i(\theta_i) = [r + \gamma\argmax_{a^{\prime}}Q(s^{\prime},a^{\prime};\theta_{i-N}) - Q(s,a;\theta_i)]\nabla_{\theta_i}Q(s,a;\theta_i)
\end{gather*}

\subsection{Challenges of RL}
In their survey work,~\citet{komer_2013} outline several ``curses'' that create challenges in the application of RL to real-world settings such as robotics. The two that will be addressed in this work are the \textit{curse of goal specification} and the \textit{curse of real-world samples}.

The curse of goal specification outlines the significant role that the design of the reward function plays in determining the success of an algorithm. Oftentimes, an iterative process of defining the reward function, called \textit{reward shaping}, is undertaken~\citep{ng_1999, DongYunlong2020PRSf, MannionPatrick2018Rsfk}. This method can prove to be undesirable as it requires a significant amount of time, domain expertise, and is not guaranteed to result in optimal agent behavior. Alternatively, researchers have explored methods for learning the reward function itself, called \textit{inverse reinforcement learning}~\citep{abbeel_2004, abb}. In this paradigm, the agent is provided with examples of an expert performing the desired task, and it learns the implicit reward function that the expert is unknowingly maximizing. However, expert examples are not always available. Finally, the ``best'' reward function for a robotic agent may be complex or impossible to define due to the large number of degrees of freedom of the robot and the environment.

The curse of goal specification motivates the use of simple reward functions. The simplest reward function is a predicate, $\mathcal{R}: \mathcal{S} \times \mathcal{A} \rightarrow \{0,1\}$, which produces a \textit{sparse reward} that only offers informative feedback when the desired behavior is achieved. This rarely-occurring learning presents obvious issues to the performance and convergence speed of the agent.

The curse of real-world samples deals with the collection of data. Unlike simulated environments, robotic agents require interaction with the physical world, which results in comparatively slow episode steps. Additionally, some robotic tasks may require human interaction. For example, the researcher may need to reset the robot at the beginning state or move objects with which the robot has interacted. Thus, gathering significantly large samples of experience may be infeasible. This curse motivates the need for sample-efficient algorithms.

To help encourage efficiency during training, most modern reinforcement learning algorithms use a replay memory that stores historical tuples of transitions, $(s, a, r, s^{\prime})$, from which minibatches are drawn to update the Q-function. A significant amount of research work has gone into understanding design implications of the replay memory~\citep{mem_size, ChaHan2020PERF, RamicicMirza2020CMRM, zilli_2208}. 

Perhaps the most impactful aspect of replay memories is the composition of the experience tuple samples drawn from the memory. Simply sampling randomly can help break any harmful temporal correlations in the agent's learning process~\citep{temp_cor}. However, more intelligent methods can help the agent autonomously decide which memory tuples are ``best''. These methods usually rely on prioritizing experiences that provide a large TD error, which suggests certain agent-environment interactions are more surprising and, therefore, informative to the agent~\citep{prior, prior2}. While these methods have shown great promise, they have the clear limitation of not being applicable in sparse-reward settings where the TD error is zero an overwhelming majority of the time.

To overcome this issue, researchers have developed several methods for steering the composition of replay memory samples in sparse-reward settings. However, some methods are limited to agents that have a specific set of characteristics \citep{pher} or require a significant amount of compute overhead through the addition of an extra model~\citep{ZuoGuoyu2020Ehrl}. Perhaps the most prominent of these methods is Hindsight Experience Replay (HER)~\citep{andry_2017}.

\subsection{Hindsight Experience Replay}
HER overcomes the issues of sparse rewards by altering the memories in the replay to present the outcomes of episodes as successful even though they may not be. At the end of each episode, the agent has the option to reproduce the episode's trajectory using a goal other than the actual goal of the task. As outlined in the original paper, the simplest strategy is to change the goal, $g \rightarrow g^{\prime}$, to be the final state reached in the episode, $s_{T}$. Using this virtual goal, the original reward for the final step in the episode is changed, $r_T \rightarrow r_{T}^{\prime}$, to one regardless of whether or not the real goal is achieved. Doing so benefits the agent by filling its replay memory with experience tuples that will provide a non-zero feedback signal. 

While HER provides a useful tool for dealing with sparse rewards, it has some downfalls. For one, the vanilla algorithm provides no guarantees that sampled minibatches will contain experience tuples with non-zero rewards. This means that it may be highly likely that some learning steps will only contain non-useful experience tuples, especially for tasks with large $T$. In addition, some draws, through random chance, may contain a relatively large number of virtual-goal experience tuples as opposed to real-goal experience tuples. This may lead to the agent learning an undesirable policy that completes a virtual goal. These issues motivate a sampling strategy that grants control over minibatch makeup, as well as one that prioritizes real-goal tuples.

\section{Or Your Money Back Sampler}
To overcome the aforementioned limitations of HER, we introduce \textit{Or Your Money Back} (OYMB), a novel replay memory sampler. It is designed to guarantee that minibatches contain a controllable amount of useful experience tuples by acting as a direct interface to the agent's replay memory. OYMB provides a set of hyperparameters to control the percent of non-zero reward tuples, $\lambda$, the decay or growth rate this percent, $\delta_{\lambda}$, and the limit of the percentage, $\lambda_{min/max}$. Together, these hyperparameters give the researcher full control over the minibatch makeup.

\begin{figure}[!h]
    \centering
    \includegraphics[scale=0.75]{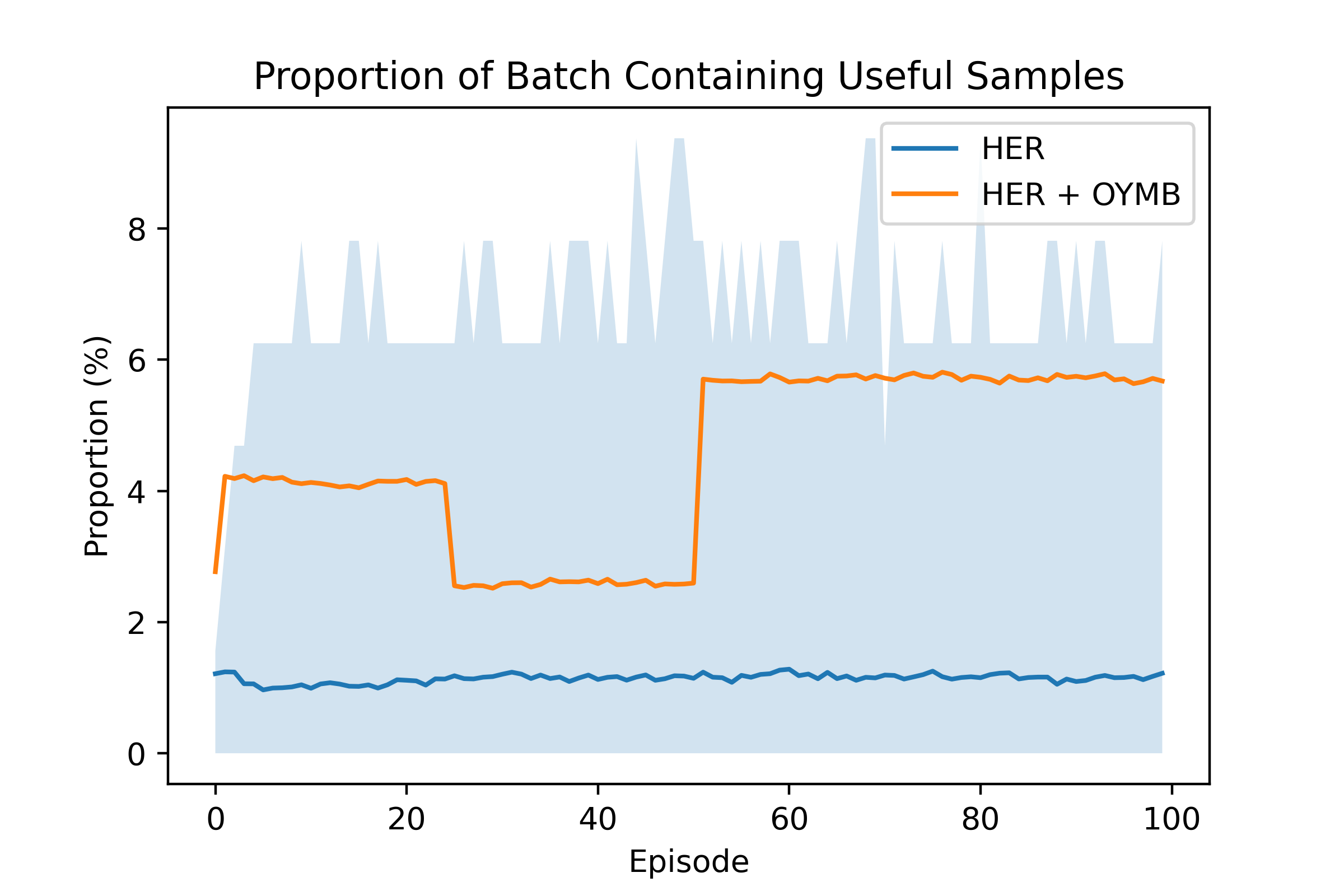}
    \caption{Proportion of non-zero samples in replay memory batch. HER + OYMB uses a manually-defined sampling schedule.}
    \label{fig:sample_pcts}
\end{figure}

For example, the researcher may set a sampling schedule in the same way they might define a schedule for the learning rate hyperparameter of a neural network. Or, the researcher may set a manually-defined schedule. Observing Figure~\ref{fig:sample_pcts}, below, we measured the proportion of non-zero reward samples in a minibatch over 100 episodes of agent training. Using OYMB, we dictated that the agent's memory provides 4\% until episode 25, 2.5\% from episodes 25 to 50, and finally 5.5\% for the remainder of training. At each episode, we drew 1000 samples from the replay memory, recording the mean, minimum, and maximum proportions. We show the means with solid lines and the spread between the minimums and maximums with shaded regions. Despite the fact that the replay memory grows as experience tuples are added, OYMB provides stable control, only showing signs of wavering from inconsistencies in floating-point precision. The proportion for vanilla HER varies wildly.
\newpage
In addition to the percentage interface, OYMB also provides a preferential-lookup scheme that prioritizes sampling experience tuples from trajectories that complete the real goal before choosing experience tuples from HER-adjusted goals. This means that, as the agent begins learning to accomplish the real goal throughout training, the minibatches will contain more real-goal tuples and less virtual-goal tuples. We accomplish this by adding two vectors to the replay memory that track the in-memory location of real-goal tuples, $\mathcal{D}.real\_indices$, and HER-altered tuples, $\mathcal{D}.HER\_indices$. Essentially, OYMB uses HER as a temporary pathway to reaching the real goal. When the agent is able to saturate its replay memory with tuples that achieve the real goal, it will no longer draw virtual goals from the memory. In tasks with dynamic goals, the core OYMB algorithm can easily be extended to guarantee virtual-goal tuples. 

The following two psuedo-code blocks outline the OYMB algorithm and then how OYMB fits within the greater training scheme.
\begin{tcolorbox}[fonttitle=\bfseries, title=OYMB Sampler]
    Algorithm parameters: $\lambda$, $\delta_{\lambda}$, $\lambda_{min/max}$, batch size $B$, replay memory $\mathcal{D}$\\
    $n \leftarrow round(B \lambda)$ \\
    $n_{real} \leftarrow min(n, \; length(\mathcal{D}.real\_indices))$ \\
    $n_{HER} \leftarrow n - n_{real}$ \\
    $n_{random} \leftarrow B - n$ \\
    \ForEach{i=1, $n_{real}$}{
        sample randomly from $\mathcal{D}.real\_indices$
    }
    \ForEach{i=1, $n_{HER}$}{
        sample randomly from $\mathcal{D}.HER\_indices$
    }
    \ForEach{i=1 in $n_{random}$}{
        sample randomly from $\mathcal{D}$
    }
    \textbf{According to schedule do} \\
    $\lambda \leftarrow \left\{ 
        \begin{array}{ll}
        \lambda_{min/max} \;\; \text{if} \;\; \delta_{\lambda}\lambda \;\; \text{outside of} \;\; \lambda_{min/max} \\
        \delta_{\lambda}\lambda \;\; \text{otherwise}
        \end{array}
    \right.$
\end{tcolorbox}

\begin{tcolorbox}[fonttitle=\bfseries, title=HER + OYMB sampler]
    Algorithm parameters: $M$, $T$, real goal $g$, batch size $B$\\
    Initialize replay memory $\mathcal{D}$, DQN $\theta$, target DQN $\theta_{target}$\\
    \ForEach{episode=1, M}{
        Observe initial $s$ \\
        \ForEach{step=1, T}{
            $a \leftarrow \pi(s||g)$ according to $\epsilon$-greedy \tcp*[f]{$||$ denotes concatenation} \\
            Perform $a$ and observe $s^{\prime}$ \\
            Store $(s||g,a,r,s^{\prime}||g)$ in $\mathcal{D}$ \\
            Sample $B$ transitions $(s||g,a,r,s^{\prime}||g)$ from $\mathcal{D}$ according to OYMB \\
            Perform optimization step on $\theta$
        }
        $g^{\prime} \leftarrow s_{T}$ \\
        \ForEach{step, $i$, in episode}{
            \uIf{$s_{i} = g^{\prime}$}{
                $r_i \leftarrow 1$ \\
            \uIf{$g^{\prime} = g$}{
                $\mathcal{D}.real\_indices||i$
            }
            \Else{
                $\mathcal{D}.HER\_indices||i$
            }
            }
        }
    $\theta_{target} \leftarrow \theta$ \tcp*[f]{Copy weights to target network} 
}
\end{tcolorbox}

\section{Experiments}
\subsection{Environments}
To test OYMB, we deployed it in three environments. The first is the discrete control version of the LunarLander environment\footnote{https://gym.openai.com/envs/LunarLander-v2/}. This task is meant to simulate the simple physics problem of gently lowering a vehicle down to a landing pad. The action space is discrete and provides four choices: left thrust, right thrust, bottom thrust, or do nothing. The state space is a vector of length eight that contains information on the lander's coordinates, velocities, angle, and indication for ground contact.

The second is the discrete control version of the MountainCar\footnote{https://gym.openai.com/envs/MountainCar-v0/} environment introduced by~\citet{mountaincar} in his PhD thesis. The MountainCar task is to drive a vehicle up a mountain to reach the top. However, the vehicle's acceleration alone is not enough to power the car all the way up. The only way to solve this task is for the agent to reverse the vehicle up a hill and use the momentum from this small hill to get up the mountain. The action space is discrete and provides three options: accelerate left, do nothing, or accelerate right. The state space is continuous and contains values for the car's position and current velocity. 

The third is an environment of our own design, called Robo. It is a 10x10 gridworld configured to be a maze.
The agent's goal is to find its way to a specified square in the maze. The actions are: go forward, turn left, or turn right. The state representation is a vector of length two that gives the agent information on their distance from the goal, as well as the distance between them and a wall they might be facing, acting like a LIDAR. The distance-to-goal measurement disregards walls, as to not pass any path information to the agent. Also, artificial noise is added to the LIDAR reading by uniformly sampling from a given range of numbers dependent on the number of tiles between the agent and the wall it is facing. For more information on the noise, see Table~\ref{tab:lidar_noise}, below. Altogether, this environment requires the agent to learn orientation behavior, as well as understand navigation in a noisy environment.
\begin{table}
    \centering
    \begin{tabular}{c c}
        Number of squares away & LIDAR reading range \\
        \toprule
         1& [10, 30] \\
         2 & [31, 80] \\
         3 & [81, 150] \\
         $\geq$4 & [151, 300] \\
         \bottomrule
    \end{tabular}
    \caption{Introducing noise into LIDAR readings}
    \label{tab:lidar_noise}
\end{table}

\begin{figure}
    \centering
    \includegraphics[scale=0.25]{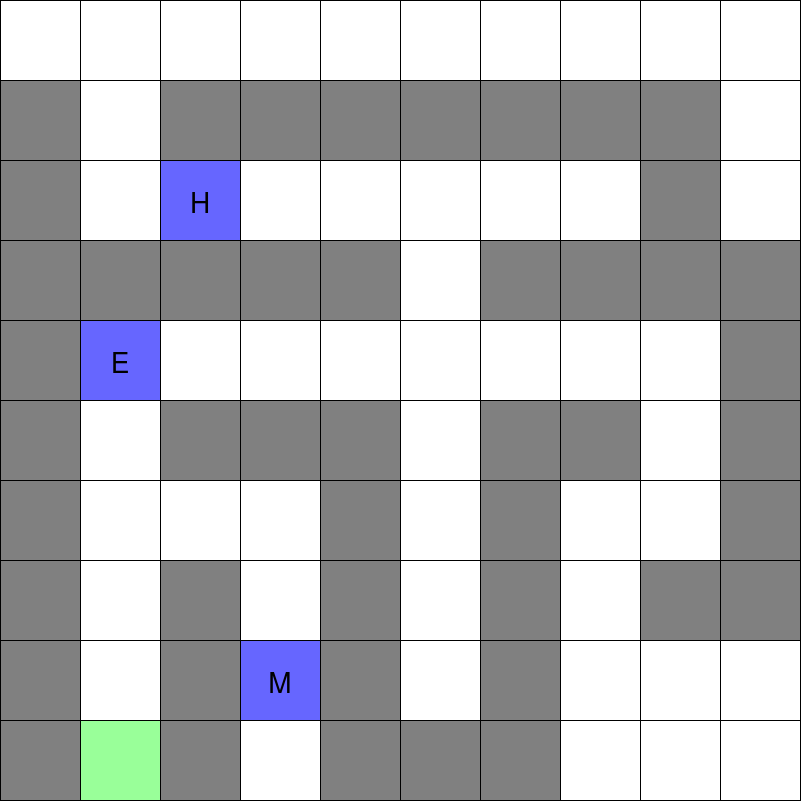}
    \caption{Depiction of the Robo environment. Dark squares are walls, the green square is the beginning position of the agent, and the purple squares denote the goal positions for the hard (H), medium (M), and easy (E) tasks.}
    \label{fig:robo_env}
\end{figure}

We evaluate our agent on three levels of difficulty in the Robo environment. The first is a ``straight shot'' walk (easy) in which the agent must simply learn to walk forward several tiles. The second is a ``U-Turn'' (medium), where the agent must learn to delay its reward by having to walk away from the goal before being able to turn around a corridor to reach the goal. The third (hard) places the goal a great distance from the agent, requiring it to traverse several corridors and avoid dead-ends. Figure~\ref{fig:robo_env} shows a graphical depiction of the Robo environment and its tasks.

\subsection{Evaluation}
For all environments, we modify the reward function to return zero for all steps that do not complete the task, otherwise return one. For each task, we ran a version of the agent with vanilla HER and then a version with HER + OYMB. The LunarLander environment has an episode length of 1,000 steps, and the MountainCar environment has an episode length of 250 steps. The Robo environment has an episode length of 150, 150, and 300 for the easy, medium, and hard task, respectively.

For all tasks, our DQNs were made of two dense layers with 64 and 32 hidden units, respectively. As done by~\citet{dqn}, we employed an $\epsilon$-greedy action-selection policy that was linearly-annealed from 1 to 0.01, dropping in value between episodes. We trained our DQNs at each episode step with a batch size of 64 using the Adam optimizer~\citep{adam}, and updated the target network's weights at the end of every episode. For the LunarLander task, we trained the agents for 1,000 episodes 15 times. For the MountainCar task, we trained the agents for 250 episodes 10 times. For each of the three tasks in the Robo environment, we trained the agents for 250 episodes 5 times. The metric used to evaluate every task is the cumulative number of successful real-goal completions.

\newpage
\section{Results}
The figures presented below depict the mean performance (solid line) and one standard deviation from the mean (shaded area) across runs. See Table~\ref{tab:hypers}, towards the end of this section, for a full description of the OYMB hyperparameters used for each task.

\begin{figure}[!h]
    \centering
    \begin{subfigure}[b]{0.49\textwidth}
        \centering
        \includegraphics[width=\textwidth]{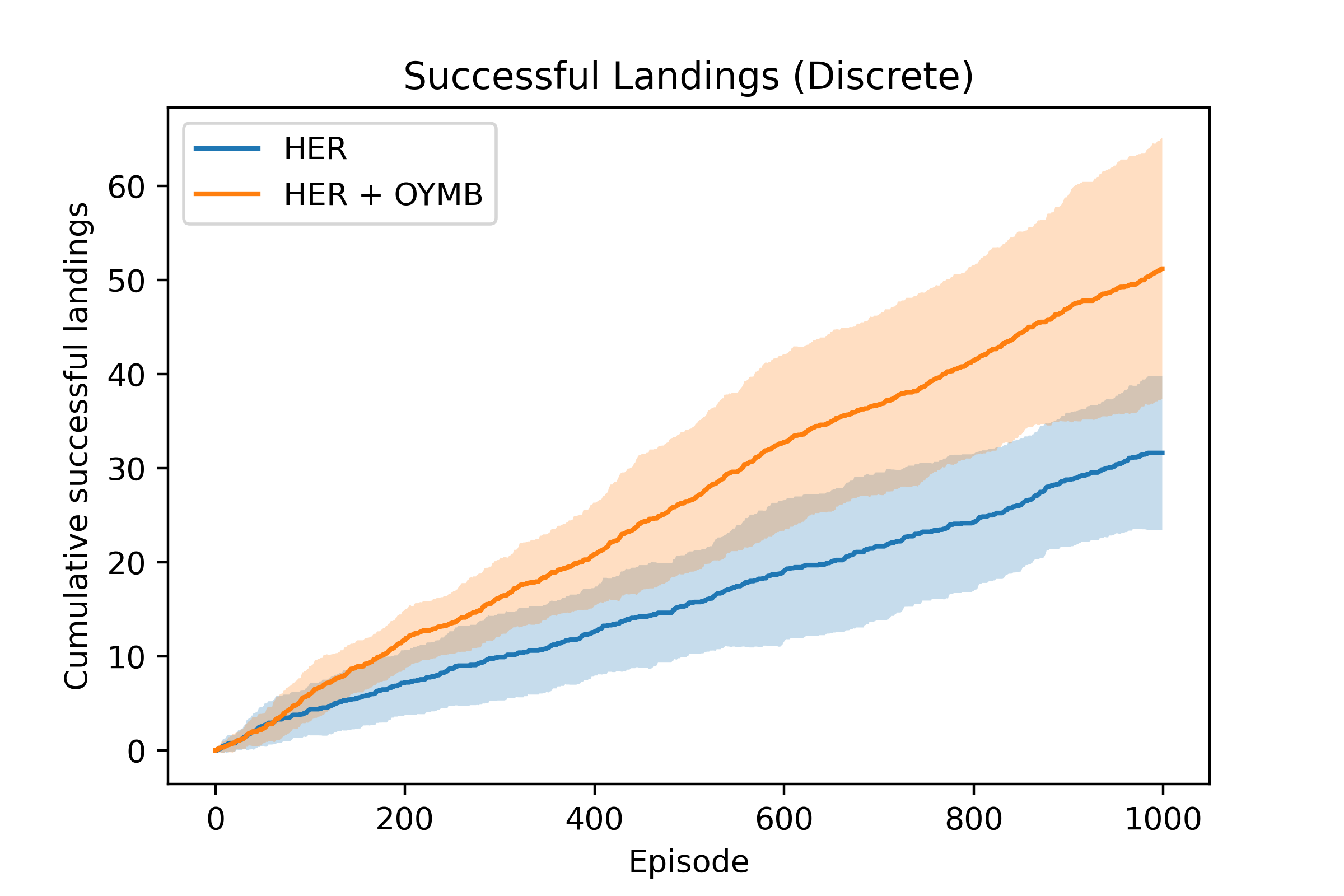}
    \end{subfigure}
    \begin{subfigure}[b]{0.49\textwidth}
        \centering
        \includegraphics[width=\textwidth]{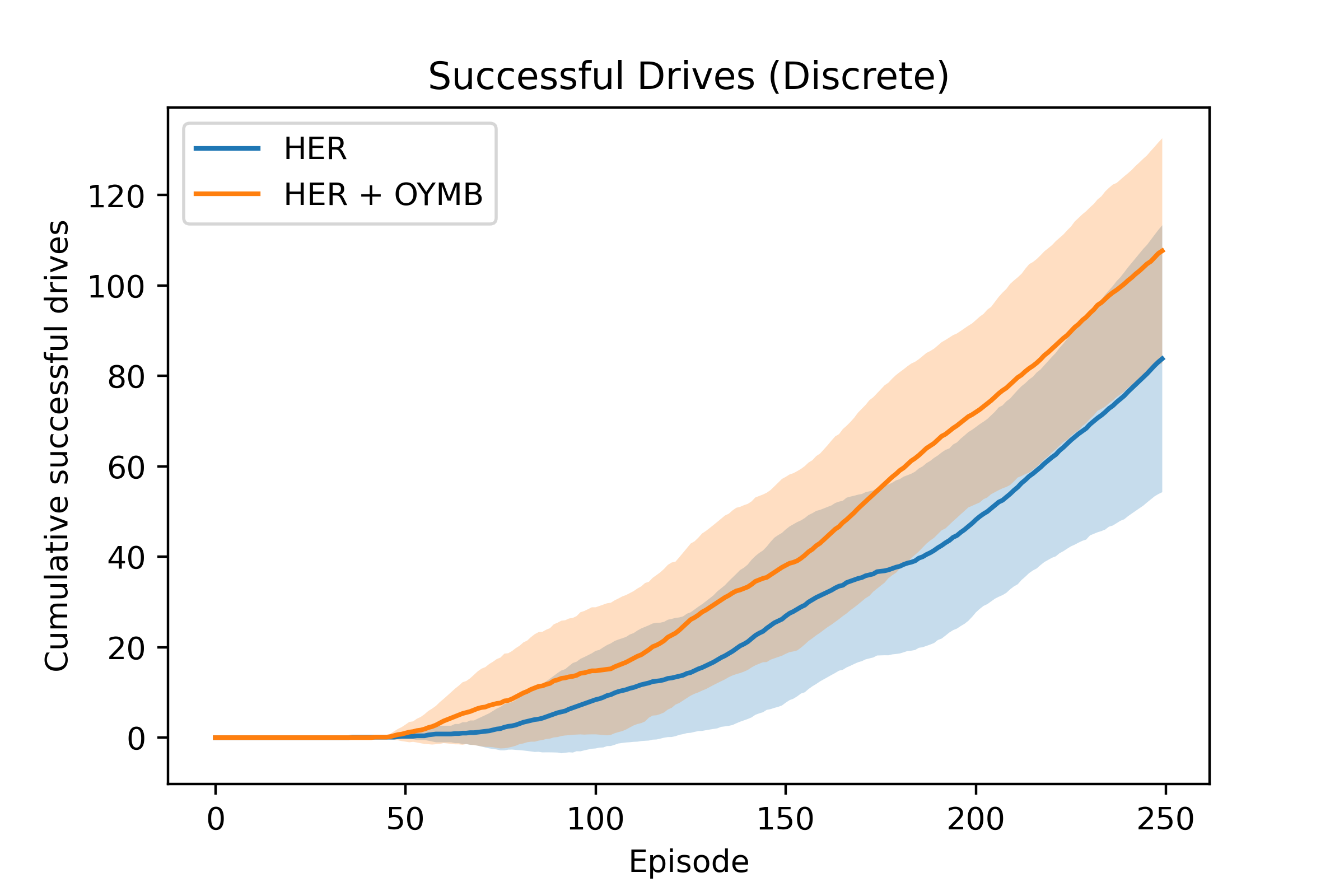}
    \end{subfigure}
    \caption{Cumulative successful episodes for sparse LunarLander (left) and sparse MaintainCar (right).}
    \label{fig:gym_results}
\end{figure}

Figure~\ref{fig:gym_results}, below, shows the results of the agents trained on the LunarLander environment (left) and the MountainCar environment (right). From both plots, we see that the agents trained with a combination of HER and OYMB were able to outperform the agents that used only HER. After 1,000 training episodes in LunarLander, the HER + OYMB agents, on average, were able to complete the real goal nearly twice as often as the HER-only agents. After 250 training episodes in MountainCar, the HER + OYMB agents were able to perform at least as well as the HER-only agents. 

\begin{figure}[!h]
    \begin{subfigure}{0.49\textwidth}
        \centering
        \includegraphics[width=\textwidth]{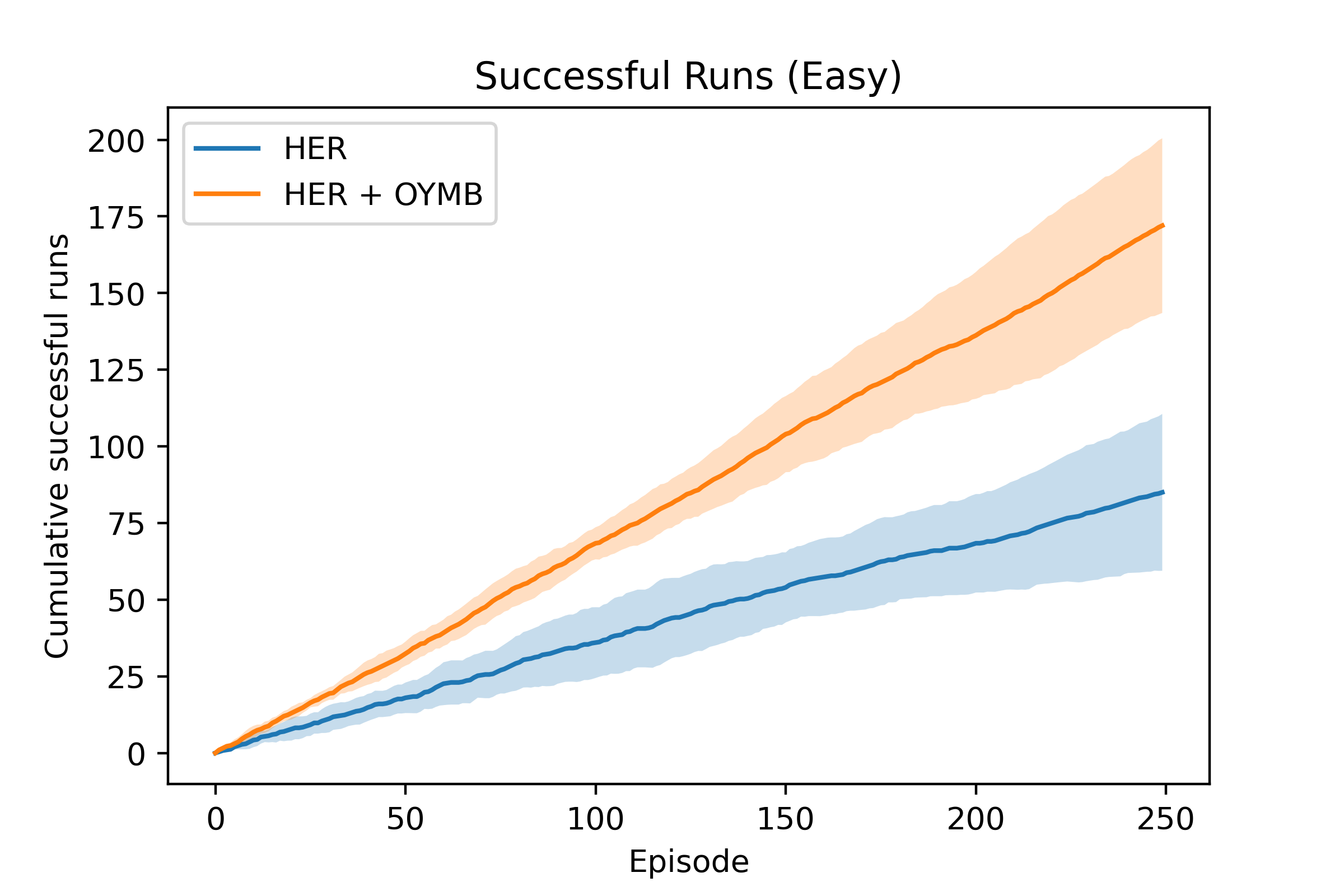}
    \end{subfigure}
    \begin{subfigure}{0.49\textwidth}
        \centering
        \includegraphics[width=\textwidth]{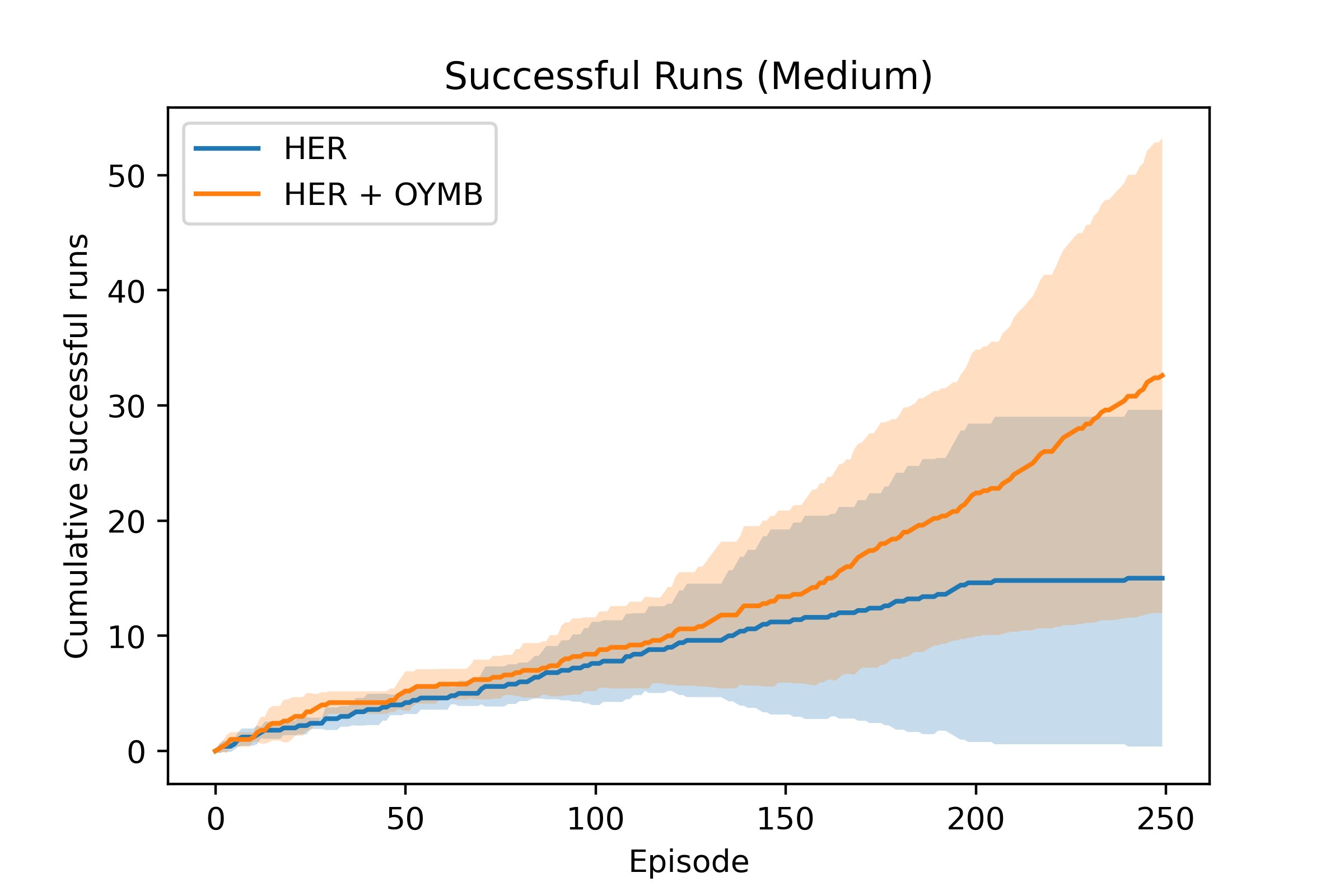}
    \end{subfigure}
    \begin{subfigure}{\textwidth}
        \centering
        \includegraphics[width=0.49\textwidth]{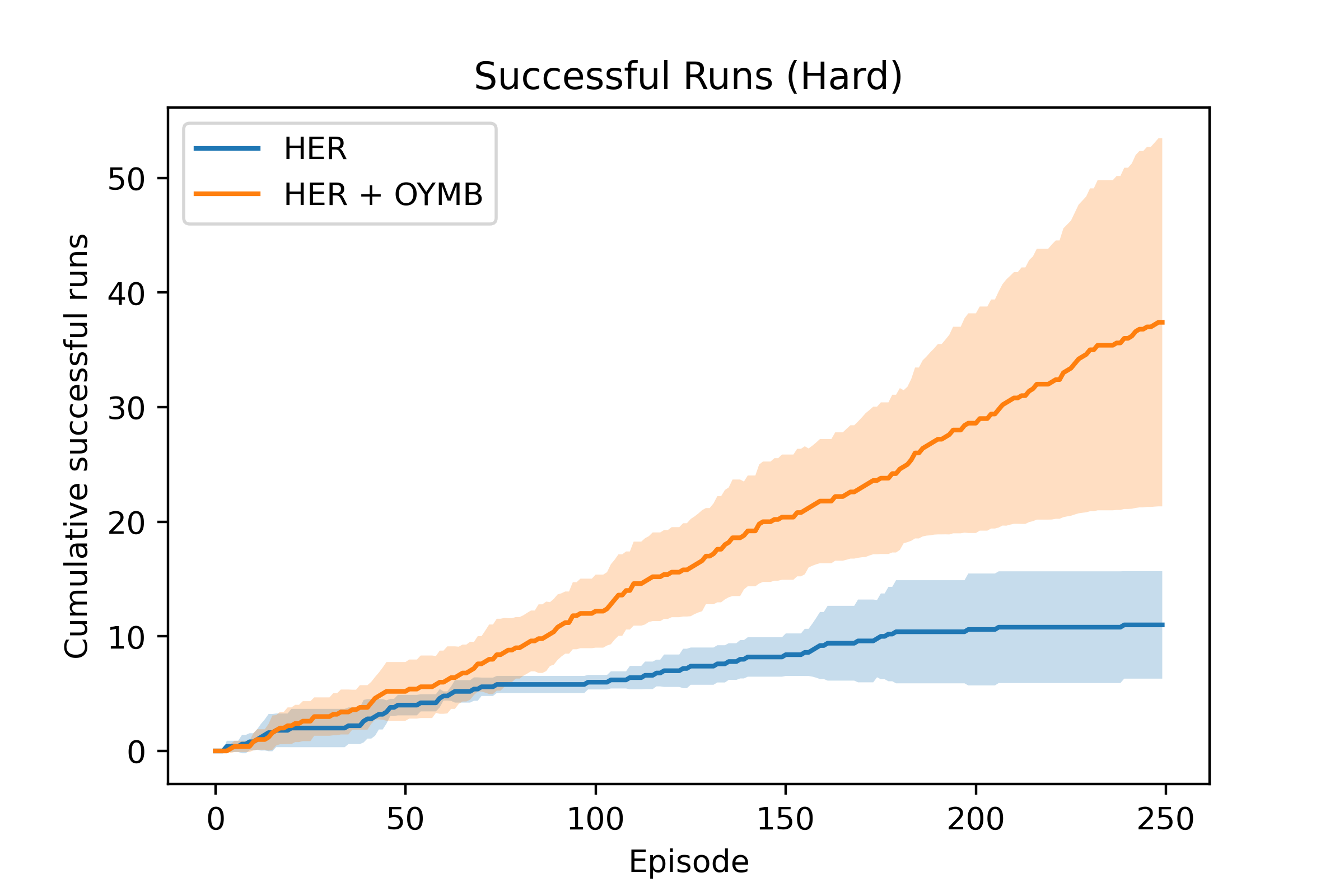}
    \end{subfigure}
    \caption{Cumulative successful episodes for the easy task (top left), medium task (top right) and hard task (bottom) of the Robo environment.}
    \label{fig:robo_results}
\end{figure}

Figure~\ref{fig:robo_results}, below, depicts the results of the three tasks in the Robo environment. In the easy task, the HER + OYMB agents were able to complete the real goal, on average, over twice as often as the HER-only agents. In the medium task, the HER + OYMB agents performed at least as well as the HER-only agents. In the hard task, the HER + OYMB agents were able to complete the real goal, on average, over four times as often as the HER-only agents. In addition, the HER-only agents showed divergent behavior in the medium and hard tasks, as represented by the marginally-decreasing trend of their cumulative successful runs metric. We did not observe this phenomenon in the HER + OYMB agents, which instead showed a trend of marginally-increasing successful runs across all three tasks.
\newline
\begin{table}[!h]
    \centering
    \begin{tabular}{cccc}
         Environment & $\lambda$ & $\delta_{\lambda}$ & $\lambda_{min/max}$ \\
         \toprule
         LunarLander & 0.65 & 0.996 & 0.01\\
         MountainCar & 0.05 & 1 & 0.05 \\
         Robo (easy) & 0.25 & 1 & 0.25\\
         Robo (medium) & 0.25 & 1 & 0.25\\
         Robo (hard) &  0.25 & 1 & 0.25\\
         \bottomrule
    \end{tabular}
    \caption{Tuned hyperparameters of OYMB for each task.}
    \label{tab:hypers}
\end{table}

\newpage

\section{Conclusion}
In this paper, we introduced \textit{Or Your Money Back} (OYMB), a replay memory sampler designed to work with the Hindsight Experience Replay (HER) memory in sparse reinforcement learning settings. OYMB acts as a direct interface to the agent's replay memory and guarantees stable control over the makeup of memory samples. Having this control allows for tuning of the minibatches that are used during the training process. In addition, OYMB uses a preferential-lookup scheme that prioritizes drawing real-goal tuples before drawing HER-adjusted tuples. We proved, empirically, across five tasks in three unique environments, that OYMB can be tuned to allow for more efficient training, resulting in better performance by the agent in a smaller number of training episodes.

\subsection{Future Work}
Currently, it is not clear how best to set the hyperparamaters of the OYMB sampler. In addition, from our experiments, it appears that the best settings may vary across environments. We hypothesize that there may be some fundamental guiding principles for optimal minibatch makeup in sparse reinforcement learning settings.

Also, for tasks with dynamic or multiple goals, it is not clear how best to balance the preferential-lookup scheme within OYMB. In tasks with multiple goals, how should we split the sampling of real-goal tuples across the goals? In tasks with dynamic goals, how can we continue to encourage exploration such that the agent does not become ``stuck'' learning from memories that accomplish a goal that no longer applies?

\bibliographystyle{apalike}  
\bibliography{references} 

\begin{thebibliography}{}

\bibitem[Abbeel, 2008]{abb}
Abbeel, P. (2008).
\newblock {\em Apprenticeship Learning and Reinforcement Learning with
  Application to Robotic Control}.
\newblock PhD thesis, Stanford University.

\bibitem[Abbeel and Ng, 2004]{abbeel_2004}
Abbeel, P. and Ng, A.~Y. (2004).
\newblock Apprenticeship learning via inverse reinforcement learning.
\newblock In {\em Proceedings of the Twenty-First International Conference on
  Machine Learning}, page~1, New York, NY, USA. Association for Computing
  Machinery.

\bibitem[Andrychowicz et~al., 2017]{andry_2017}
Andrychowicz, M., Wolski, F., Ray, A., Schneider, J., Fong, R., Welinder, P.,
  McGrew, B., Tobin, J., Abbeel, P., and Zaremba, W. (2017).
\newblock Hindsight experience replay.
\newblock In {\em 31st Conference on Neural Information Processing Systems},
  Long Beach, CA. ACM.

\bibitem[Cha et~al., 2020]{ChaHan2020PERF}
Cha, H., Park, J., Kim, H., Bennis, M., and Kim, S.-L. (2020).
\newblock Proxy experience replay: Federated distillation for distributed
  reinforcement learning.
\newblock {\em IEEE Intelligent Systems}, pages 1--1.

\bibitem[Dong et~al., 2020]{DongYunlong2020PRSf}
Dong, Y., Tang, X., and Yuan, Y. (2020).
\newblock Principled reward shaping for reinforcement learning via lyapunov
  stability theory.
\newblock {\em Neurocomputing (Amsterdam)}, 393:83--90.

\bibitem[Horgan et~al., 2018]{prior2}
Horgan, D., Quan, J., Budden, D., Barth-Maron, G., Hessel, M., van Hasselt, H.,
  and Silver, D. (2018).
\newblock Distributed prioritized experience replay.
\newblock In {\em ICLR 2018}, Vancouver, BC, Canada.

\bibitem[Kingma and Ba, 2014]{adam}
Kingma, D. and Ba, J. (2014).
\newblock Adam: A method for stochastic optimization.

\bibitem[Kober et~al., 2013]{komer_2013}
Kober, J., Bagnell, J.~A., and Peters, J. (2013).
\newblock Reinforcement learning in robotics: A survey.
\newblock {\em The International Journal of Robotics Research},
  32(11):1238--1274.

\bibitem[Liu and Zou, 2017]{mem_size}
Liu, R. and Zou, J. (2017).
\newblock The effects of memory replay in reinforcement learning.

\bibitem[Mannion et~al., 2018]{MannionPatrick2018Rsfk}
Mannion, P., Devlin, S., Duggan, J., and Howley, E. (2018).
\newblock Reward shaping for knowledge-based multi-objective multi-agent
  reinforcement learning.
\newblock {\em Knowledge engineering review}, 33.

\bibitem[Mnih et~al., 2015]{dqn}
Mnih, V., Kavukcuoglu, K., Silver, D., Graves, A., Antonoglou, I., Wierstra,
  D., and Riedmiller, M. (2015).
\newblock Human-level control through deep reinforcement learning.
\newblock {\em Nature}, 518.

\bibitem[Moore, 1990]{mountaincar}
Moore, A. (1990).
\newblock {\em Efficient Memory-Based Learning for Robot Control}.
\newblock PhD thesis, University of Cambridge.

\bibitem[Ng et~al., 1999]{ng_1999}
Ng, A., Harada, D., and Russell, S. (1999).
\newblock Policy invariance under reward transformations: Theory and
  application to reward shaping.
\newblock In {\em Proceedings of the Sixteenth International Conference on
  Machine Learning}, page 278–287, San Francisco, CA, USA. Morgan Kaufmann
  Publishers Inc.

\bibitem[Novati and Koumoutsakos, 2019]{temp_cor}
Novati, G. and Koumoutsakos (2019).
\newblock Remember and forget for experience replay.
\newblock In {\em Proceedings of the 36th International Conference on Machine
  Learning}, Long Beach, California. PMLR.

\bibitem[Ramicic and Bonarini, 2020]{RamicicMirza2020CMRM}
Ramicic, M. and Bonarini, A. (2020).
\newblock Correlation minimizing replay memory in temporal-difference
  reinforcement learning.
\newblock {\em Neurocomputing (Amsterdam)}, 393:91--100.

\bibitem[Ren and Ben-Tzvi, 2020]{RenHailin2020Arlt}
Ren, H. and Ben-Tzvi, P. (2020).
\newblock Advising reinforcement learning toward scaling agents in continuous
  control environments with sparse rewards.
\newblock {\em Engineering Applications of Artificial Intelligence}, 90.

\bibitem[Schaul et~al., 2016]{prior}
Schaul, T., Quan, J., Antonglou, I., and Silver, D. (2016).
\newblock Prioritized experience replay.

\bibitem[Seo et~al., 2019]{SeoMinah2019RPCA}
Seo, M., Vecchietti, L.~F., Lee, S., and Har, D. (2019).
\newblock Rewards prediction-based credit assignment for reinforcement learning
  with sparse binary rewards.
\newblock {\em IEEE Access}, 7:118776--118791.

\bibitem[Sutton and Barto, 2018]{sutton_barto_2018}
Sutton, R.~S. and Barto, A.~G. (2018).
\newblock {\em Reinforcement Learning: An Introduction}.
\newblock The MIT Press, Cambridge, MA.

\bibitem[Wen and Van~Roy, 2017]{WenZheng2017ERLi}
Wen, Z. and Van~Roy, B. (2017).
\newblock Efficient reinforcement learning in deterministic systems with value
  function generalization.
\newblock {\em Mathematics of Operations Research}, 42(3):762--782.

\bibitem[Yang et~al., 2020]{YangDujia2020SERL}
Yang, D., Qin, X., Xu, X., Li, C., and Wei, G. (2020).
\newblock Sample efficient reinforcement learning method via high efficient
  episodic memory.
\newblock {\em IEEE Access}, 8:1--1.

\bibitem[Zhao and Tresp, 2018]{pher}
Zhao, R. and Tresp, V. (2018).
\newblock Energy-based hindsight experience prioritization.
\newblock In {\em Conference on Robot Learning}, Zurich, Switzerland.

\bibitem[Zilli and Hasselmo, 2008]{zilli_2208}
Zilli, E. and Hasselmo, M. (2008).
\newblock The influence of markov decision process structure on the possible
  strategic use of working memory and episodic memory.
\newblock {\em PLoS One}, 3(7).

\bibitem[Zuo et~al., 2020]{ZuoGuoyu2020Ehrl}
Zuo, G., Zhao, Q., Lu, J., and Li, J. (2020).
\newblock Efficient hindsight reinforcement learning using demonstrations for
  robotic tasks with sparse rewards.
\newblock {\em International Journal of Advanced Robotic Systems}, 17(1).

\end{thebibliography}

\end{document}